%% file: neurips_2026.tex
\documentclass{article}

\usepackage[dblblindworkshop, final]{neurips_2026}

\usepackage{marvosym}
\usepackage{wrapfig}
\usepackage[most]{tcolorbox}
\usepackage{caption}
\usepackage[utf8]{inputenc} 
\usepackage[T1]{fontenc}    
\usepackage{hyperref}       
\usepackage{url}            
\usepackage{booktabs}       
\usepackage{amsfonts}       
\usepackage{nicefrac}       
\usepackage{microtype}      
\usepackage{xcolor}         


\title{A Citation-Grounded Benchmark for Trustworthy Earnings Call Transcript Analysis with Large Language Models}
\workshoptitle{Third NeurIPS Workshop on Attributing Model Behavior at Scale: Data Attribution and Provenance}

\author{
	Yingzhu Zhao\thanks{These authors contributed equally to this work.},
	Vlad Pandelea\footnotemark[1],
	Han Yuan\footnotemark[1],
	Bo Hu\footnotemark[1],
	\\
	{\bf Wuqiong Luo\footnotemark[1],}
{\bf Li Zhang,}
	{\bf Zheng Ma}\thanks{Correspondence: Zheng Ma, Singapore Decision Science Center
		of Excellence, American Express, 1 Marina Boulevard, 018989, Singapore.}
	\\
	Global Decision Science, American Express
	\\
	\texttt{\{Yingzhu.Zhao, Vlad.A.Pandelea, Han.Yuan1, Bo.Hu\}@aexp.com}
	\\
	\texttt{\{Wuqiong.Luo, Li.Zhang1, Zheng.Ma2\}@aexp.com}
}

\begin{document}

\maketitle

\begin{abstract}
Large language models (LLMs) have been increasingly used for financial document analysis, including earnings call transcripts (ECTs). Beyond generating standalone claims, users increasingly prefer grounded analyses that pair claims with verifiable citations from source documents to enable independent validation. However, evaluating such analytical claims typically requires extensive expert annotation, which is costly and difficult to scale, and real-world financial analysis commonly involves long context-question-answer triplets, further increasing task complexity. To address these challenges and benchmark the current landscape of grounded analysis by LLMs, we propose a numeric evidence evaluation method that enables groundedness assessment without reliance on expert annotation. We also introduce an automated dataset construction pipeline and construct ECTs-100 from the top 100 constituents of the S\&P 500 to support benchmark of both groundedness and correctness. In addition, we examine conscious incompetence, a practical failure mode in financial analysis in which LLMs must detect when available evidence is insufficient and refrain from producing unsupported hallucinations. Empirical results show that LLMs perform well in groundedness but face notable limitations in correctness, with informational insufficiency presenting an additional challenge.
\end{abstract}

\section{Introduction}
Earnings call transcripts (ECTs) are textual records of conference calls that capture discussions between company executives and investors regarding financial performance, question-answer (QA) sessions, and future outlooks of publicly traded firms \citep{SubjECTive}. Due to the jargon-heavy and content-dense nature, ECTs present significant challenges for efficiently analyzing unstructured content and identifying informative insights \cite{huang-etal-2021-template,chen-takamura-2024-term}. 

The development of large language models (LLMs) has made automated financial analysis of ECTs feasible. Beyond generating standalone claims, modern LLMs can produce grounded answers that pair claims with verifiable citations from source documents, enabling independent validation and increasing user trust. However, evaluating claim-citation pairs remains challenging. Although financial professionals such as investment analysts can systematically interpret earnings call transcripts and produce well structured analytical claims, such expert annotations are costly and difficult to scale \citep{SubjECTive}. LLMs-based alternatives are inexpensive and scalable but suffer from model-dependent preference \citep{liu-etal-2023-g}. Moreover, real-world analysis of ECTs often involves long context-question-answer triplets, further increasing task complexity \citep{hu2025extract}.

To address these challenges and systematically benchmark grounded answers in the form of claim citation pairs, we propose a numeric evidence evaluation method that measures alignment of financially salient numbers across claims, citations, and source documents. Given the central role of quantitative information in financial analysis \citep{hu-etal-2025-fintrust}, this approach serves as a proxy for groundedness without requiring costly expert annotations. Using this method, we benchmark common grounded generation baselines on expert-designed long questions over ECTs from the top 100 constituents of the S\&P 500 index. We further introduce a fully automated dataset construction pipeline and construct ECTs-100, which provides context question answer triplets and facilitates evaluation of both correctness and groundedness. Finally, we examine conscious incompetence, a failure mode in financial analysis in which LLMs must detect when available evidence is insufficient and refrain from hallucinations \citep{li-etal-2024-towards-verifiable}. Empirical results on grounded financial analysis indicate that modern LLMs achieve strong groundedness, suggesting that they rarely hallucinate citations absent from source documents. However, further evaluation on ECTs-100 reveals persistent limitations in correctness: although the generated claim citation pairs are grounded in the provided context, they often fail to answer the question accurately, with informational insufficiency presenting an additional challenge. We hope that our findings, along with the proposed numeric evidence evaluation and automated dataset construction pipeline, will accelerate trustworthy financial analysis.

\section{Methods}
\subsection{Preliminaries}
Before presenting the numeric evidence evaluation, automated dataset construction pipeline, and evaluation metrics, we define two central notions used throughout this study: \emph{groundedness} and \emph{correctness}. Groundedness measures whether a generated claim or citation is explicitly supported by evidence in the provided context, regardless of whether it fully answers the target question. Correctness is stricter: it measures whether the generated answer identifies the information required by the question and whether the cited evidence supports that required information. Thus, a correct answer should be grounded, whereas a grounded answer need not be correct.
Let $Q$ denote a question and let $C$ denote the source context. In this study, $C$ corresponds to the earnings call transcript documents. A model-generated answer is denoted by $\widetilde{A}=(\widetilde{A}^{\mathrm{clm}},\widetilde{A}^{\mathrm{cit}})$, where $\widetilde{A}^{\mathrm{clm}}$ denotes the generated claims and $\widetilde{A}^{\mathrm{cit}}$ denotes the citations provided by the model. We use $\widetilde{A}^{\mathrm{clm}}{i}$ to denote the $i$-th generated claim, where $i=1,\ldots,\widetilde{I}$ and $\widetilde{I}$ is the number of generated claims. For each generated claim $\widetilde{A}^{\mathrm{clm}}{i}$, we use $\widetilde{A}^{\mathrm{cit}}{i,k}$ to denote its $k$-th associated citation segment, where $k=1,\ldots,\widetilde{K}{i}$ and $\widetilde{K}_{i}$ is the number of citation segments attached to generated claim $i$.
This work focuses on numeric evidence, specifically salient financial figures. We use $F^{\mathrm{ctx}}{u}$ to denote the $u$-th salient numeric figure extracted from the context $C$, where $u=1,\ldots,U$. The full set of context figures is denoted by $\mathcal{F}^{\mathrm{ctx}}(C)$. For generated claim $\widetilde{A}^{\mathrm{clm}}{i}$, let $\widetilde{\mathcal{F}}^{\mathrm{clm}}{i}$ denote the set of extracted numeric figures $\widetilde{F}^{\mathrm{clm}}{i,j}$, where $j=1,\ldots,\widetilde{J}{i}$ and $\widetilde{J}{i}$ is the number of extracted claim figures. For generated citation segment $\widetilde{A}^{\mathrm{cit}}{i,k}$, let $\widetilde{\mathcal{F}}^{\mathrm{cit}}{i,k}$ denote the set of extracted numeric figures $\widetilde{F}^{\mathrm{cit}}{i,k,l}$, where $l=1,\ldots,\widetilde{L}{i,k}$ and $\widetilde{L}_{i,k}$ is the number of extracted citation figures.
For correctness evaluation, which relies on reference answers, we use non-tilded notation. The ground-truth answer is denoted by $A=(A^{\mathrm{clm}},A^{\mathrm{cit}})$. We use $A^{\mathrm{clm}}{i}$ to denote the $i$-th ground-truth claim, where $i=1,\ldots,I$ and $I$ is the number of ground-truth claims. For each ground-truth claim $A^{\mathrm{clm}}{i}$, we use $A^{\mathrm{cit}}{i,k}$ to denote its $k$-th ground-truth citation segment, where $k=1,\ldots,K_i$. Let $\mathcal{F}^{\mathrm{clm}}{i}$ denote the set of numeric figures $F^{\mathrm{clm}}{i,j}$ extracted from ground-truth claim $A^{\mathrm{clm}}{i}$, where $j=1,\ldots,J_i$. Let $\mathcal{F}^{\mathrm{cit}}{i,k}$ denote the set of numeric figures $F^{\mathrm{cit}}{i,k,l}$ extracted from ground-truth citation segment $A^{\mathrm{cit}}{i,k}$, where $l=1,\ldots,L{i,k}$. Throughout the paper, context and source documents refer to the same object $C$.

\subsection{Numeric evidence evaluation}
Conventional benchmarks for grounded generation typically rely on expert human annotation, which is costly and difficult to scale in long-context question-answering settings. LLMs-based alternatives are cheaper and more scalable, but they may exhibit preferences toward LLMs-generated text \citep{liu-etal-2023-g,yuan2025toward}. To address these limitations, we propose \emph{numeric evidence evaluation} (NEE), an automatic evaluation procedure that compares salient numeric information across generated claims, generated citations, ground-truth answers, and source contexts.

Following the notation in the preliminaries, NEE operates on numeric evidence extracted from four sources: context figures $\mathcal{F}^{\mathrm{ctx}}(C)$, generated-claim figures $\widetilde{\mathcal{F}}^{\mathrm{clm}}{i}$, generated-citation figures $\widetilde{\mathcal{F}}^{\mathrm{cit}}{i,k}$, and, when correctness evaluation is required, ground-truth figures $\mathcal{F}^{\mathrm{clm}}{i}$ and $\mathcal{F}^{\mathrm{cit}}{i,k}$. These evidence sets are obtained using structured output \citep{structured-output}. Each extracted numeric item, whether it is $F^{\mathrm{ctx}}{u}$, $\widetilde{F}^{\mathrm{clm}}{i,j}$, $\widetilde{F}^{\mathrm{cit}}{i,k,l}$, $F^{\mathrm{clm}}{i,j}$, or $F^{\mathrm{cit}}_{i,k,l}$, follows the same tuple format $F=(s,v,t,m)$. Here, $s$ is the text span expressing the numeric fact, $v$ is either a scalar value or a two-endpoint interval, $t\in{\mathrm{scalar},\mathrm{range}}$ indicates whether the item is a scalar or range expression, and $m$ is an optional explicit unit.
The extraction prompt instructs the model to identify only explicit measurable quantities and to exclude numbers that do not express substantive measurements. We distinguish measurement-bearing numbers from incidental numeric tokens because many numbers in financial documents are not factual quantities relevant to answer grounding. Dates, fiscal periods, section labels, product names, model identifiers, and version numbers are excluded because their numeric components typically serve as identifiers or temporal labels rather than measurable claims. Including such tokens would inflate numeric overlap without improving factual support. Range expressions are preserved as single atomic numeric claims rather than split into separate scalar claims. A phrase such as ``between 46\% and 47\%'' expresses one interval-valued assertion, not two independent assertions about 46\% and 47\%. Treating ranges as single units also prevents partial credit when evidence contains only one endpoint of an interval. Thus, a range-valued numeric item is considered matched only when the comparison text contains the corresponding range-valued item.
After extraction, each numeric item is normalized into a comparison key $\kappa(F)=(t,\nu(v),\mu(m))$, where $\nu(\cdot)$ normalizes scalar values or interval endpoints and $\mu(\cdot)$ normalizes units. Two numeric items $F$ and $F'$ are treated as an exact numeric match, denoted by $F\equiv F'$, if $\kappa(F)=\kappa(F')$. For any two numeric evidence sets $\mathcal{X}$ and $\mathcal{Y}$, we define $\mathcal{X}\preceq\mathcal{Y}$ as the subset of items in $\mathcal{X}$ that are matched by at least one item in $\mathcal{Y}$ under $\equiv$. If there is no numeric content to verify, we define the score to be $1$.

\begin{table*}[]
	\centering
	\resizebox{\textwidth}{!}{%
		\begin{tabular}{ccccccc}
			\hline
			Baseline & Vanilla-index & Vanilla-text & PostCite-index & PostCite-text & LLMR-index & LLMR-text \\ \hline
			Claim-context & 0.975 & 0.970 & 0.975 & 0.982 & 0.977 & 0.982 \\
			Citation-context & 0.995 & 0.989 & 0.995 & 0.994 & 0.993 & 0.993 \\
			Claim-citation & 0.928 & 0.917 & 0.907 & 0.952 & 0.927 & 0.959 \\ \hline
		\end{tabular}%
	}
	\caption{Comparison across claim-context, citation-context, and claim-citation groundedness}
	\label{tab:groundedness}
\end{table*}

\subsection{Automated dataset construction}
While NEE enables groundedness evaluation of citation-grounded generation, it does not by itself assess whether generated answers identify the correct information required by the question. To support correctness benchmarking in the long-triplet setting for financial analysis, we develop an automated dataset construction pipeline that derives ground-truth claims and citations from salient financial figures.
Given a context $C$, we first extract the set of salient context figures $\mathcal{F}^{\mathrm{ctx}}(C)$. We use $F^{\mathrm{ctx}}{u}$ to denote the $u$-th extracted context figure, where $u=1,\ldots,U$. For each context figure $F^{\mathrm{ctx}}{u}$, we prompt GPT-4.1 to generate a sub-question $Q_u$ whose answer is the figure itself. The ground-truth claim for $Q_u$ is denoted by $A^{\mathrm{clm}}{u}$ and contains the target numeric figure $F^{\mathrm{ctx}}{u}$. Its corresponding numeric evidence set is therefore $\mathcal{F}^{\mathrm{clm}}{u}$, which contains $F^{\mathrm{ctx}}{u}$ as the target figure.
The sentence in $C$ containing $F^{\mathrm{ctx}}{u}$ is used as the ground-truth citation segment $A^{\mathrm{cit}}{u,1}$ \citep{song-etal-2024-veriscore}. The numeric evidence set of this citation is denoted by $\mathcal{F}^{\mathrm{cit}}{u,1}$ and includes the salient figure $F^{\mathrm{ctx}}{u}$, along with any other salient figures appearing in the same citation sentence. This process yields figure-level ground-truth triplets $(Q_u,A^{\mathrm{clm}}{u},A^{\mathrm{cit}}{u,1})$, where $u=1,\ldots,U$. In the notation of the preliminaries, each figure-level triplet corresponds to one ground-truth claim index $i=u$ and one associated citation segment, so $I=U$ and $K_i=1$ for all ground-truth claims in the constructed dataset. We keep each figure-level question-answer pair as an independent evaluation instance. Specifically, each context figure $F^{\mathrm{ctx}}_{u}$ induces one instance $(Q_u,A^{\mathrm{clm}}_{u},A^{\mathrm{cit}}_{u,1})$, where $Q_u$ is the generated question, $A^{\mathrm{clm}}_{u}$ is the corresponding ground-truth answer, and $A^{\mathrm{cit}}_{u,1}$ is the sentence-level ground-truth citation. This design preserves the long-context setting because each question is still answered with respect to the full source context $C$. This construction follows the intuition that each salient figure induces a verifiable atomic question-answer pair, while the concatenated question-answer format evaluates whether models can handle financial long triplets \citep{tan-etal-2024-proxyqa}. 

\subsection{Evaluation metrics}
\subsubsection{Groundedness}
\label{section groundedness metrics}

We evaluate groundedness at three complementary levels: (1) \emph{claim--citation groundedness}, measuring whether generated claims are supported by their associated citations; (2) \emph{claim--context groundedness}, measuring whether claims are supported by the source context independently of the generated citations; and (3) \emph{citation--context groundedness}, measuring whether the generated citations are faithful to the source context. Following prior work \citep{wu-etal-2025-natural,liu2026semisupervised}, we quantify each relationship using the proportion of numeric figures that achieve normalized exact matches. This hierarchical evaluation captures both citation coverage and numerical hallucination, providing a more fine-grained assessment than prior approaches that consider only one or two of these relationships \citep{liu-etal-2023-evaluating,li-etal-2024-towards-verifiable}. Detailed metric definitions are provided in Appendix~\ref{app:metrics}.

\subsubsection{Correctness}

Groundedness alone does not assess whether the model provides the information required by the question. We therefore evaluate \emph{claim correctness} by comparing numeric figures in generated answers with those in the ground-truth answers, using precision, recall, and F1 based on normalized exact matching. We evaluate \emph{citation correctness} by comparing the identifiers of generated citations with those of the ground-truth evidence, without requiring textual exact matching. Metrics are first averaged over QA instances within each document, and we report the mean and standard error across documents. Full definitions and aggregation procedures are provided in Appendix~\ref{app:metrics}. Together, these metrics distinguish whether an output is \emph{grounded in its evidence} from whether it is \emph{correct and relevant to the question}, enabling a comprehensive evaluation of both claims and their citations.

\section{Experiments}
This section begins by outlining the baseline methods used to generate citation-grounded analyses. Although other approaches exist, the methods selected here are representative, conceptually straightforward, and our benchmark are easily extended to other baselines. We subsequently conduct a benchmark study of groundedness evaluation methods and select a balanced method for use in subsequent correctness evaluation. After that, we assess the selected method in a real-world setting of conscious incompetence, which challenges LLMs to recognize informational insufficiency.

\subsection{Baseline methods}
\label{section baseline methods}
Following \citep{sun-etal-2024-towards-verifiable}, we evaluate three baseline paradigms: Vanilla \citep{nakano2021webgpt,gao-etal-2023-enabling}, PostCite \citep{gao-etal-2023-rarr}, and LLMs-based retrieval (LLMR) \citep{luo-etal-2024-large} under two citation formats of index and text, yielding six baselines in total. Vanilla requires LLMs to jointly generate claims and citations. PostCite decouples generation by first producing claims and then identifying the most relevant supporting sentences as citations. LLMR reverses this order, requiring the model to first select relevant sentences as citations and subsequently generate claims based solely on the selected citations. For citation formats, index assigns a unique identifier to each sentence in the source ECTs. Models output claims in text form and citations as sentence identifiers, which are then mapped back to the original transcripts. In contrast, text omits identifier preprocessing and requires models to generate both claims and citations directly as text. All baselines are implemented using GPT-4.1 under a zero-shot setting to minimize confounding effects and ensure methodological consistency. Unless otherwise specified, we adopt the default model parameters.

In summary, all six baseline methods generate a variable number of claims in response to each question, with each claim accompanied by one or more original sentences from the ECTs as citations. This structure emulates the format commonly used by financial analysts \citep{structured-output}.

\subsection{Groundedness benchmark}
Building on the baseline methods described above, we benchmark citation-grounded analysis by LLMs from the standpoint of groundedness. We first detail the setup and then report empirical results.

\subsubsection{Benchmark setup}
\label{Benchmark setup}
We collect fourth-quarter 2024 ECTs disclosed by the publicly traded top 100 constituents of the S\&P 500 index.
Because Berkshire Hathaway does not hold regular earnings calls, we substitute it with The Cigna Group, the 101st constituent in the index. A complete list of the companies and corresponding ECTs is available online.\footnote{\href{https://github.com/Han-Yuan-Med/citation-grounded-financial-analysis/tree/main/ECTs-2024-Q4}{GitHub link.}} ECTs typically consist of two main sections: prepared remarks and Q\&A session. In the prepared remarks, key company executives, such as the CEO or CFO, provide a detailed overview of the company’s financial performance, highlight major business developments, and offer insights into future prospects. The Q\&A session involves unscripted interactions between the management team and financial analysts or investors, covering a wide range of topics including strategic initiatives, the competitive landscape, and forward-looking guidance. Given that both sections are critical for comprehensive financial analysis, we include the full transcript as context. 

For the groundedness benchmark, we construct five long questions in Appendix \ref{appendix_question}, covering five thematic dimensions: (1) discussion focus and outlook, (2) risk assessment, (3) strategic growth and long-term plan, (4) macroeconomic impact, and (5) financial and capital allocation management.



\begin{wrapfigure}{r}{0.4\columnwidth}
	\vspace{-10pt}
	\centering
	\includegraphics[width=\linewidth]{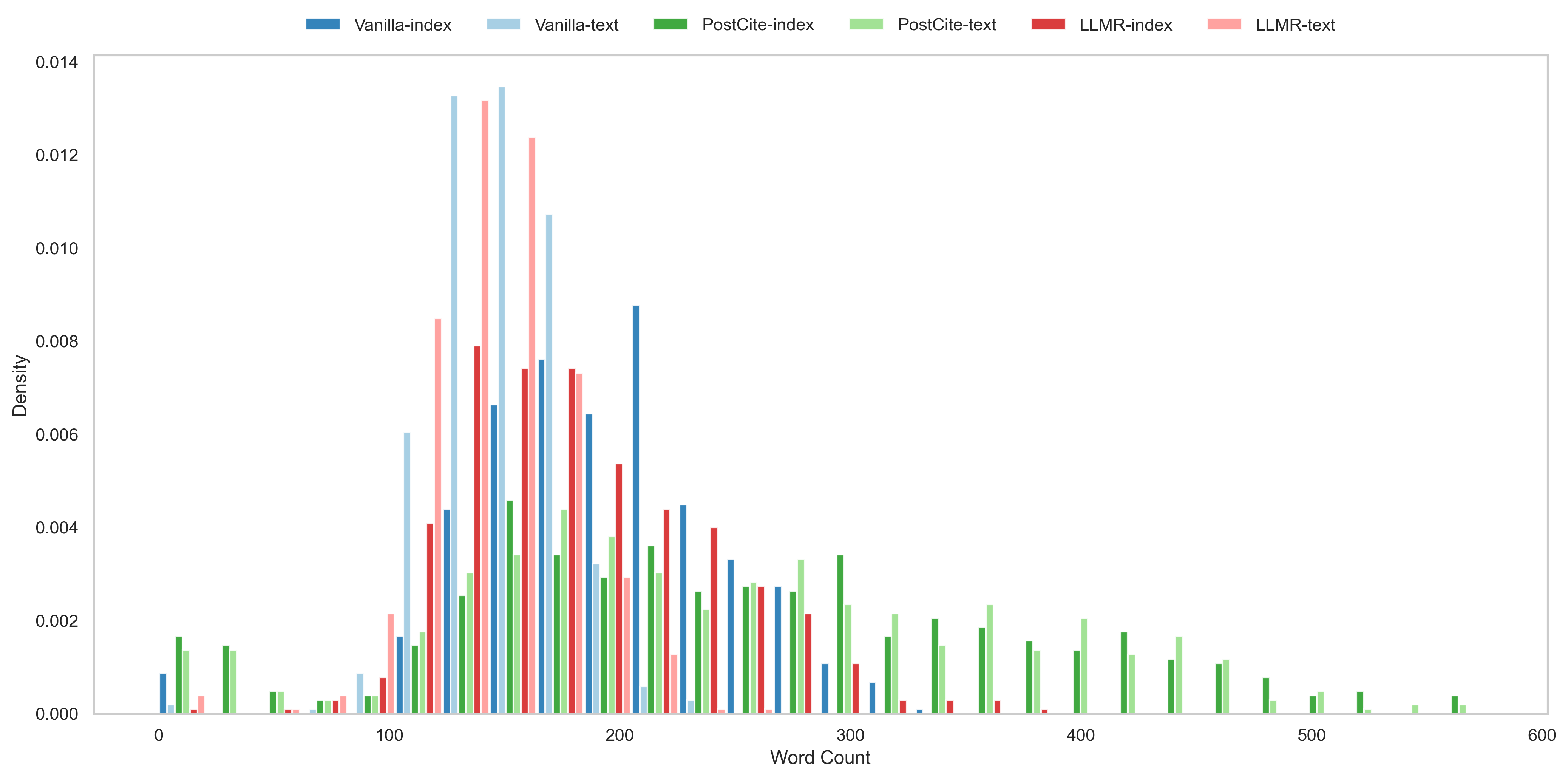}
	\caption{Claims word distribution.}
	\label{fig:density}
	\vspace{-20pt}
\end{wrapfigure}

\subsubsection{Benchmark results}
We implement the six baselines and Figure~\ref{fig:density} shows the density distributions of generated claim lengths across the six methods. The distributions are concentrated around 100-300 words, indicating a long-claim regime. Together with the long context of earnings call transcripts and expert-designed long-form questions, this pattern characterizes our long context-question-answer triplets.

Table~\ref{tab:groundedness} reports the groundedness metrics of \emph{claim--citation}, \emph{claim--context}, \emph{citation--context} detailed in Section~\ref{section groundedness metrics} for the six baseline methods detailed in Section~\ref{section baseline methods}. Overall, all methods achieve high \emph{claim--context} and \emph{citation--context groundedness}, showing that most generated numeric claims and citations are supported by the source transcript. The main difference appears in \emph{claim--citation groundedness}, which evaluates whether the cited evidence directly supports the generated claim. Although LLMR-text obtains the highest \emph{claim--citation groundedness}, LLMR-index achieves comparable groundedness, with a \emph{claim--context} score of 0.977, \emph{citation--context} score of 0.993, and \emph{claim--citation} score of 0.927. More importantly, its citation index is useful for downstream citation correctness evaluation, where we need to match generated citation evidence to the corresponding ground-truth citation. \citet{li-etal-2024-towards-verifiable} also report strong alignment between citations and claims in modern LLMs. However, their evaluation relies on natural language inference to determine whether a generated claim is entailed by a generated citation, which can be affected by the high semantic similarity inherent in LLMs-generated text \citep{wu-etal-2025-natural}. In contrast, our NEE leverages objective numeric evidence.

\subsection{Correctness benchmark}
Beyond groundedness, which evaluates whether generated claims or citations are explicitly supported by the source documents, we build ECTs-100 to enable an evaluation on correctness for citation-grounded generations. Each sample in ECTs-100 consists of a context-question-answer triplet, where the answer contains ground-truth claims and citations. This design allows us to evaluate not only the groundedness of model outputs by comparing claims or citations against the source documents but also the correctness of the generated claims and citations by comparing them against the ground-truth. We first describe the construction of ECTs-100 using the proposed automated pipeline and then present the correctness evaluation results.

\subsubsection{ECTs-100}
Consistent with the groundedness benchmark, we construct ECTs-100 using the same ECTs as source documents. In contrast, we employ LLMs to generate questions that target specific financial figures to ensure that the resulting claims and citations remain objective and free from subjective interpretation \citep{tan-etal-2024-proxyqa}. Given the large volume of financial figures in ECTs, we restrict our analysis to salient figures that appear in at least one citation in the former groundedness benchmark. The concrete question-answer-citation is provided here.\footnote{\href{https://github.com/Han-Yuan-Med/citation-grounded-financial-analysis/tree/main/ECTs-100}{GitHub link.}}



\begin{wrapfigure}{r}{0.4\columnwidth}
	\vspace{-10pt}
	\centering
	\includegraphics[width=\linewidth]{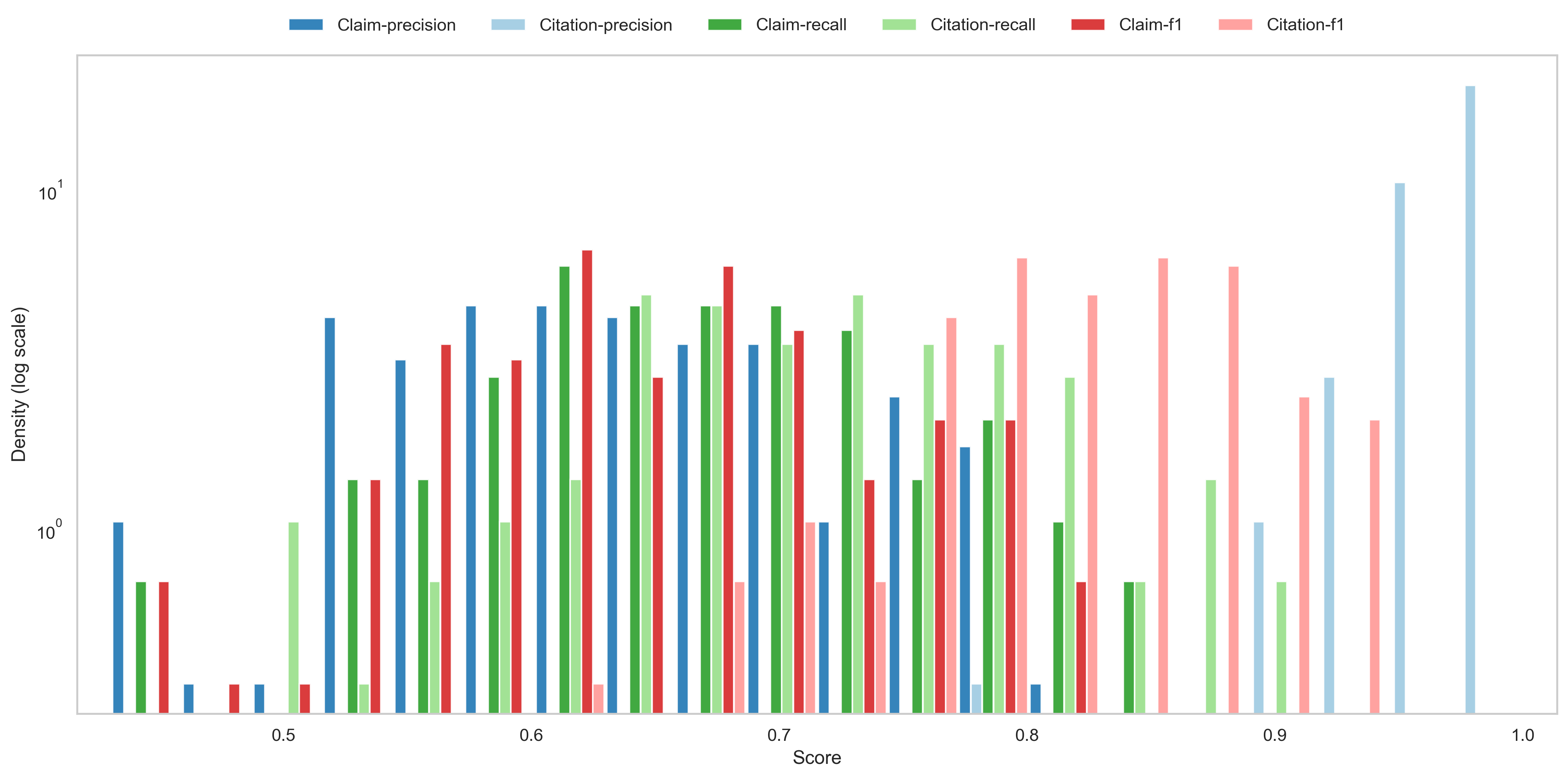}
	\caption{Correctness distribution.}
	\label{fig:scoredensity}
	\vspace{-10pt}
\end{wrapfigure}

\subsubsection{Benchmark results}
Across ECTs-100, citation precision is substantially higher than the other metrics, reaching 0.9721 (0.0032), where values are reported as mean (SE). This indicates that, when the model provides a citation, the cited evidence usually matches the correct reference. However, citation recall is much lower at 0.7139 (0.0087), leading to a citation F1 score of 0.8204 (0.0063). This gap suggests that the main limitation is citation coverage rather than citation accuracy. For claim-level metrics, recall slightly exceeds precision. Normalized claim recall reaches 0.6658 (0.0079), while normalized claim precision is 0.6296 (0.0083), yielding a normalized claim F1 score of 0.6468 (0.0080). This indicates that the model recovers a reasonable portion of the claim, but also introduces some unmatched content.

The distributional results in Figure~\ref{fig:scoredensity} are consistent with these aggregate metrics. Citation precision is tightly concentrated near the upper bound, while citation recall and citation F1 are more dispersed and centered at lower values. Claim-level precision, recall, and F1 are concentrated in the mid-score range, roughly between 0.55 and 0.75, with larger variance than citation precision. Overall, the results show that citations are reliable when they are produced, but claim correctness should be improved.

\begin{wrapfigure}{r}{0.4\columnwidth}
	\vspace{-10pt}
	\centering
	\includegraphics[width=\linewidth]{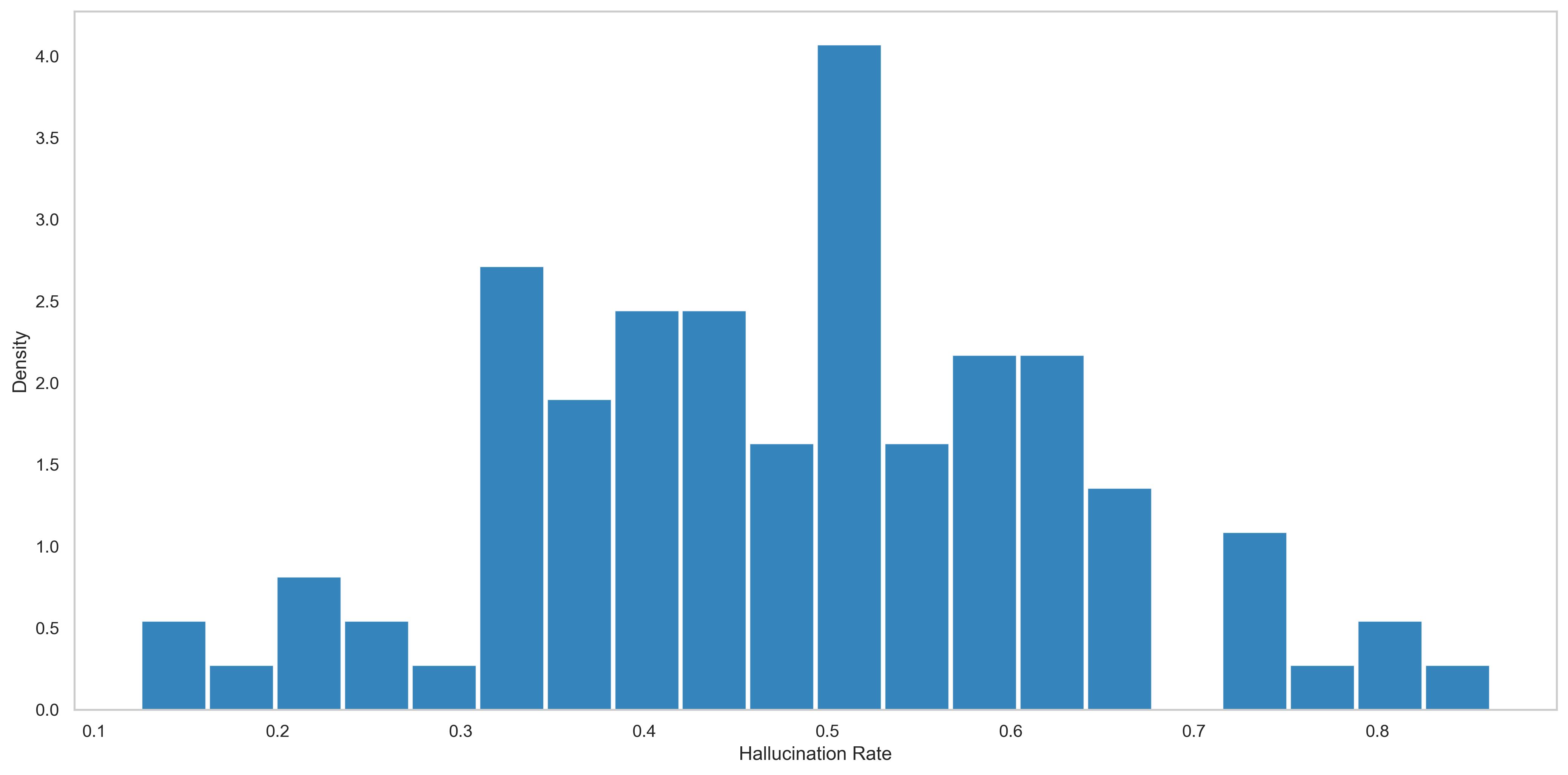}
	\caption{Hallucination rate distribution under conscious incompetence.}
	\label{fig:hallucination}
	\vspace{-20pt}
\end{wrapfigure}

\subsection{Conscious incompetence}
The former experiment examines a self-contained setting in which questions can be answered solely using the provided context. In contrast, real-world financial analysis requires LLMs to recognize when the available context is insufficient and to abstain from answering rather than producing unsupported hallucinations. Building on ECTs-100, we simulate this conscious incompetence scenario to stress-test LLMs' citation-grounded analytical capability. 

Specifically, for each document in ECTs-100, we remove the ground-truth citations containing the target figures from the context for each question, while keeping the question and all remaining context unchanged. Since the supporting evidence is no longer available, LLMs should abstain from answering and state that the question cannot be answered based on the provided document, instead of hallucinating figures from irrelevant context or fabricating figures absent from the document. We report hallucination behavior under the conscious incompetence setting. For a given question, we consider the output hallucinated if the model generates answer claims containing salient numbers or provides citations to support those claims, despite the question being informationally insufficient. Across all documents, the average hallucination rate is 0.4834 (0.0149). Figure~\ref{fig:hallucination} shows the distribution of document-level hallucination rates. The hallucination rate indicates that informational insufficiency presents an additional challenge for LLMs in ECT analysis: rather than abstaining, models can produce numerically specific claims or  citations even when the required context is unavailable.

\section{Conclusion}
In this study, we benchmark LLMs' citation-grounded analysis on earnings call transcript analysis. We introduce a numeric evidence evaluation method without the requirement of expert annotation in groundedness assessment. We further develop an automated dataset construction pipeline and develop ECTs-100 to enable further evaluation of correctness. Empirical results demonstrate that LLMs are generally reliable in maintaining groundedness, but they face challenges in preserving correctness over long context-question-answer triplets and in handling informational insufficiency.

\section{Disclaimer}
This paper is provided solely for informational purposes as an academic contribution by the authors to the research community and does not represent, reflect, or constitute the views, policies, positions, or practices of American Express or its affiliates. Nothing in this paper should be cited or relied upon as evidence of, or support for, the business views, policies, positions, or practices of American Express or its affiliates.

\bibliographystyle{unsrtnat}
\bibliography{custom}


\appendix

\section{Limitations}
\label{sec:limitation}
Our experiments demonstrate the grounded financial analytical capabilities of LLMs under long context-question-answer settings. While our study is based on earnings call transcripts, which already present a long-context challenge with an average length of 9373.84 words, real-world financial analysis often involves even longer documents, such as Form 10-Q and Form 10-K filings, which typically exceed 25,000 words \citep{koval-etal-2024-learning}. As earnings calls generally take place shortly after the release of these financial reports, future work will explore combining ECTs with the corresponding 10-Q or 10-K documents to construct more comprehensive contexts and facilitate more insightful analysis \citep{rekabsaz-etal-2017-volatility}. Also, we acknowledge that the automated dataset construction cannot fully replace datasets developed through expert panel discussions led by financial professionals. However, we adopt it as a practical proxy with several advantages: targeting specific financial figures ensures that the resulting claims and citations remain objective and largely free from subjective interpretation, while the automated pipeline enables low-cost dataset construction and rapid evaluation of foundation models' financial capabilities. Additionally, NEE should be interpreted as a scalable proxy for numeric evidence alignment rather than a complete measure of semantic entailment or a replacement for expert assessment. Because numeric matching is based on normalized numeric type, value, and unit, identical figures may be counted as matches even when they refer to different entities, financial concepts, or reporting periods, and support for non-numeric portions of a claim is not directly evaluated. We also do not validate NEE against human or expert groundedness judgments in this study, and therefore do not claim that its scores are interchangeable with human assessments. Moreover, all six baseline configurations are implemented using GPT-4.1, so the reported results characterize different citation-generation strategies under a single underlying model rather than performance across LLM families. Finally, because ECTs-100 is constructed automatically, with GPT-4.1 used to generate questions from salient source figures, its reference answers and citations should be interpreted as automatically constructed references.

\section{Ethical considerations}
\label{sec:dataset_ethics}
ECTs-100 is constructed entirely from publicly available earnings call transcripts and does not include any internal or proprietary information. We acknowledge that financial analysis can be inherently subjective; therefore, questions and answers in ECTs-100 focus on salient and objective financial figures and are designed to reduce subjective interpretation. This dataset is intended solely for research purposes, and users are encouraged to ensure that it aligns with their specific requirements.

\section{Related work}
\subsection{ECTs analysis}
ECTs from publicly listed companies offer rich insights for financial analysis and have catalyzed the development of natural language processing (NLP) techniques in the finance domain \citep{choe-etal-2023-exploring,anderson-etal-2024-greenback}. \citet{huang-etal-2021-template} construct a set of candidate templates and employ Longformer \citep{beltagy2020longformer} to generate summaries of ECTs based on the selected template. Similarly, \citet{mukherjee-etal-2022-ectsum} develop dataset of ECTSum, which consists of ECTs paired with bullet-point-style summaries extracted from Reuters. In contrast, our ECTs-100 offers a more fine-grained format by designing specific QA pairs, rather than simply presenting a comprehensive list of points covering various aspects of financial analysis \citep{goldsack-etal-2025-facts}. Also, each answer in our dataset comprises both claim and citation. \citet{an-etal-2024-l} identify the unique challenge posed by the long-context nature of ECTs and develop open-ended QA pairs based on them. Our dataset differs in two key aspects: we incorporate long-form questions and answers to constitute long triplets, whereas their work primarily focuses on short and simple queries such as “What is the restructuring plan?”; ECTs-100 comprehensively covers constituents from diverse industries, compared to their limited 6 companies. These enhancements underscore the distinctive value of ECTs-100.

In addition to analytical text generation, ECTs have been used in various financial analytical tasks that output either continuous or categorical values such as stock price \citep{handschke-etal-2018-corpus,keith-stent-2019-modeling,sang-bao-2022-dialoguegat,niu-etal-2023-kefvp,koval-etal-2023-forecasting,yang-etal-2023-measuring,zhou-he-2023-causal,shah-etal-2024-numerical,lu-etal-2025-strux,zhang-etal-2025-evaluating}, earnings surprise \citep{koval-etal-2024-financial}, business risk type \citep{leidner-schilder-2010-hunting}, strategy maturity \citep{pataci-etal-2022-digicall}, financial subjectivity \citep{SubjECTive,takayanagi-etal-2025-gpt}, and argument relation  \citep{otiefy-alhamzeh-2024-exploring}. Also, ECTs have also been utilized in tasks unrelated to financial analysis, including automatic speech recognition \citep{mathur-etal-2023-personalm} and vagueness classification \citep{theil-stuckenschmidt-2020-predicting}.

\subsection{Grounded analysis}
LLMs are known to produce coherent and persuasive outputs that may nonetheless contain factual inaccuracies \citep{chen-etal-2025-graphcheck}, which can lead to monetary losses in financial analysis \citep{yuan2025exploring}. Consequently, grounded analysis that includes both claims and explicit citations is preferable to the default paradigm of claim-only generation, particularly in long-context settings \citep{honovich-etal-2022-true-evaluating}. Early studies on LLMs have recognized the advantages of citation-grounded generation and trained models to answer questions while providing supporting evidence \citep{nakano2021webgpt,gao-etal-2023-enabling}. As LLMs have evolved, their foundational capabilities have improved to the point that they can now generate citation-grounded responses without task-specific fine-tuning. Nevertheless, evaluations of generative search engines in late 2023 revealed that only 51.5\% of generated claims were supported by their accompanying citations, based on human assessment \citep{liu-etal-2023-evaluating}. Over the past two years, citation-grounded generations have continued to advance. For example, \citet{asai2024self} and \citet{sun-etal-2024-towards-verifiable} instruct LLMs to reflect on retrieved information and selectively incorporate relevant evidence to produce grounded responses within a retrieval-augmented generation (RAG) framework. \citet{sun-etal-2024-towards-verifiable} combine the Vanilla and PostCite methods by first generating claims and citations jointly through RAG and then refining the citations based on the initial outputs. More recent studies further show that LLMs can outperform conventional retrievers and be directly employed as retrievers to identify relevant evidence \citep{luo-etal-2024-large}.

\section{Evaluation metrics}
\label{app:metrics}
\subsection{Groundedness}
\label{section groundedness metrics}
We evaluate groundedness at three complementary levels. First, \emph{claim--citation groundedness} measures whether each generated claim is supported by the citations provided for that claim. Second, \emph{claim--context groundedness} measures whether generated claims are supported by the source context, independent of the model-provided citations. Third, \emph{citation--context groundedness} measures whether the generated citations themselves are faithful to the source context.
For \emph{claim--citation groundedness}, let $\widetilde{\mathcal{F}}^{\mathrm{cit}}{i,*}$ denote the set of all numeric figures appearing in the citation segments associated with generated claim $\widetilde{A}^{\mathrm{clm}}{i}$. This set is obtained by pooling $\widetilde{\mathcal{F}}^{\mathrm{cit}}{i,k}$ over $k=1,\ldots,\widetilde{K}{i}$. 
we measure the proportion of numeric figures in generated claims that are supported by their associated citations: $\frac{\sum{i=1}^{\widetilde{I}}|\widetilde{\mathcal{F}}^{\mathrm{clm}}{i}\preceq\widetilde{\mathcal{F}}^{\mathrm{cit}}{i,*}|}{\sum_{i=1}^{\widetilde{I}}|\widetilde{\mathcal{F}}^{\mathrm{clm}}_{i}|}$. This metric evaluates whether the numeric claims made by the model are covered by the citations it provides.
For \emph{claim--context groundedness}, we compute the proportion of numeric figures in generated claims that can be matched to the source context: $\frac{\sum{i=1}^{\widetilde{I}}|\widetilde{\mathcal{F}}^{\mathrm{clm}}{i}\preceq\mathcal{F}^{\mathrm{ctx}}(C)|}{\sum{i=1}^{\widetilde{I}}|\widetilde{\mathcal{F}}^{\mathrm{clm}}{i}|}$. For \emph{citation--context groundedness}, we compute the proportion of numeric figures in generated citations that can be matched to the source context: $\frac{\sum_{i=1}^{\widetilde{I}}\sum_{k=1}^{\widetilde{K}{i}}|\widetilde{\mathcal{F}}^{\mathrm{cit}}{i,k}\preceq\mathcal{F}^{\mathrm{ctx}}(C)|}{\sum_{i=1}^{\widetilde{I}}\sum_{k=1}^{\widetilde{K}{i}}|\widetilde{\mathcal{F}}^{\mathrm{cit}}{i,k}|}$. These exact-match ratios quantify numeric hallucination in generated claims and citations, respectively \citep{wu-etal-2025-natural}. Compared with prior work that typically evaluates only one or two of these relationships \citep{liu-etal-2023-evaluating,li-etal-2024-towards-verifiable}, our benchmark provides a more fine-grained and hierarchical assessment of citation-grounded financial analysis.

\subsection{Correctness}
Groundedness evaluates whether generated content is supported by the source context, but it does not determine whether the model has selected the information required by the question. Since our automated pipeline provides ground-truth answers and citations, we evaluate correctness by comparing model outputs against these references.

We use $d$ to index source documents, where $d=1,\ldots,D$. For document $d$, let $n=1,\ldots,N_d$ index its QA instances. For the $n$-th QA instance in document $d$, let $\mathcal{F}^{\mathrm{clm}}_{d,n}$ denote the set of ground-truth answer figures, and let $\widetilde{\mathcal{F}}^{\mathrm{clm}}_{d,n}$ denote the set of figures extracted from the model answer. Since a model answer may contain multiple figures, we evaluate claim correctness using precision, recall, and F1. Instance-level claim precision $\mathrm{Prec}^{(d,n)}_{\mathrm{clm}}$ is the proportion of generated answer figures that match ground-truth figures: $\frac{|\widetilde{\mathcal{F}}^{\mathrm{clm}}_{d,n}\preceq\mathcal{F}^{\mathrm{clm}}_{d,n}|}{|\widetilde{\mathcal{F}}^{\mathrm{clm}}_{d,n}|}$. Instance-level claim recall $\mathrm{Rec}^{(d,n)}_{\mathrm{clm}}$ is the proportion of ground-truth figures recovered by the generated answer: $\frac{|\mathcal{F}^{\mathrm{clm}}_{d,n}\preceq\widetilde{\mathcal{F}}^{\mathrm{clm}}_{d,n}|}{|\mathcal{F}^{\mathrm{clm}}_{d,n}|}$. Instance-level claim $\mathrm{F1}^{(d,n)}_{\mathrm{clm}}$ is $\frac{2\mathrm{Prec}^{(d,n)}_{\mathrm{clm}}\mathrm{Rec}^{(d,n)}_{\mathrm{clm}}}{\mathrm{Prec}^{(d,n)}_{\mathrm{clm}}+\mathrm{Rec}^{(d,n)}_{\mathrm{clm}}}$, with F1 defined as $0$ when both precision and recall are $0$. The matching relation $\preceq$ follows the normalized numeric exact match defined earlier, where two numeric items $F$ and $F'$ match if $\kappa(F)=\kappa(F')$.

Document-level claim correctness $\mathrm{Prec}^{(d)}_{\mathrm{clm}}$, $\mathrm{Rec}^{(d)}_{\mathrm{clm}}$, and $\mathrm{F1}^{(d)}_{\mathrm{clm}}$ are obtained by averaging over the QA instances in each document: $\frac{1}{N_d}\sum_{n=1}^{N_d}\mathrm{Prec}^{(d,n)}_{\mathrm{clm}}$, $\frac{1}{N_d}\sum_{n=1}^{N_d}\mathrm{Rec}^{(d,n)}_{\mathrm{clm}}$, and $\frac{1}{N_d}\sum_{n=1}^{N_d}\mathrm{F1}^{(d,n)}_{\mathrm{clm}}$, respectively. Finally, we report the mean and standard error (SE) over document-level metrics. For example, claim precision is reported as $\frac{1}{D}\sum_{d=1}^{D}\mathrm{Prec}^{(d)}_{\mathrm{clm}}$ with SE $\frac{\mathrm{sd}(\{\mathrm{Prec}^{(d)}_{\mathrm{clm}}\}_{d=1}^{D})}{\sqrt{D}}$, and recall and F1 are reported analogously.

For citations, let $\mathcal{A}^{\mathrm{cit}}_{d,n}$ denote the set of ground-truth citations for QA instance $n$ in document $d$, and let $\widetilde{\mathcal{A}}^{\mathrm{cit}}_{d,n}$ denote the set of generated citations. We do not require textual exact match between citation strings. Instead, we evaluate whether the generated citations recover the ground-truth citation evidence through identifier matching, using the citation-index method described in the following sections. Instance-level citation precision $\mathrm{Prec}^{(d,n)}_{\mathrm{cit}}$ is $\frac{|\{\,\widetilde{A}\in\widetilde{\mathcal{A}}^{\mathrm{cit}}_{d,n}:\exists A\in\mathcal{A}^{\mathrm{cit}}_{d,n}\ \mathrm{s.t.}\ \mathrm{ind}(\widetilde{A})=\mathrm{ind}(A)\,\}|}{|\widetilde{\mathcal{A}}^{\mathrm{cit}}_{d,n}|}$, and instance-level citation recall $\mathrm{Rec}^{(d,n)}_{\mathrm{cit}}$ is $\frac{|\{\,A\in\mathcal{A}^{\mathrm{cit}}_{d,n}:\exists \widetilde{A}\in\widetilde{\mathcal{A}}^{\mathrm{cit}}_{d,n}\ \mathrm{s.t.}\ \mathrm{ind}(\widetilde{A})=\mathrm{ind}(A)\,\}|}{|\mathcal{A}^{\mathrm{cit}}_{d,n}|}$, where $\mathrm{ind}(\cdot)$ denotes citation-identifier extraction. Instance-level citation $\mathrm{F1}^{(d,n)}_{\mathrm{cit}}$ is $\frac{2\mathrm{Prec}^{(d,n)}_{\mathrm{cit}}\mathrm{Rec}^{(d,n)}_{\mathrm{cit}}}{\mathrm{Prec}^{(d,n)}_{\mathrm{cit}}+\mathrm{Rec}^{(d,n)}_{\mathrm{cit}}}$, with F1 defined as $0$ when both precision and recall are $0$. Document-level citation precision, recall, and F1 are computed by averaging their instance-level values over $n=1,\ldots,N_d$, and the final reported citation metrics are the mean and SE over the $D$ document-level scores.

These metrics evaluate whether the model outputs the required answers and their corresponding citations, rather than merely producing numerically grounded but irrelevant content to the question.

\section{Questions}
\label{appendix_question}
We prepare questions corresponding to the five thematic dimensions described in Section \ref{Benchmark setup}, and the detailed content for each question is presented in Figures \ref{Question on focus of discussion and outlook} to \ref{Question on financial and capital allocation management}, respectively.

\begin{tcolorbox}[enhanced]
	Identify the central themes and key topics that were the focus of the call discussion. This should include areas such as financial performance, forward guidance, strategic initiatives, operational highlights, industry trends, competitive forces and responses to external factors. Forward-looking guidance should include financial projections, revenue growth expectations, operating expenses, profitability targets, capital allocation, leverage and any strategic initiatives or market performance projections for the upcoming periods. Pay attention to how management emphasizes growth opportunities, challenges, overall strategic direction and consistency of message.
\end{tcolorbox}
\noindent\begin{minipage}{\columnwidth}
	\captionof{figure}{Question on focus of discussion and outlook}\label{Question on focus of discussion and outlook}
\end{minipage}

\begin{tcolorbox}[enhanced]
	Identify the potential risks and challenges mentioned by the company's management during the call. Focus on risks related to operational issues, market volatility, geopolitics, obsolescence, regulatory changes, competition, disruptions, significant industry trends and macroeconomic factors that could affect the business's future performance and strategic objectives. Significant industry trends should include technological advancements, changes in consumer behavior, competitive dynamics, and regulatory developments.
\end{tcolorbox}
\noindent\begin{minipage}{\columnwidth}
	\captionof{figure}{Question on risk and threat assessment}\label{Question on risk and threat assessment}
\end{minipage}

\begin{tcolorbox}[enhanced]
	Describe the key strategic initiatives and long-term plans discussed by management in the call. This should include short- and long-term goals related to market expansion, geographic expansion, organic growth, product development, cost management, partnerships, mergers and acquisitions and other growth-focused activities that support the company's overall strategy.
\end{tcolorbox}
\noindent\begin{minipage}{\columnwidth}
	\captionof{figure}{Question on strategic growth and long-term planning}\label{Question on strategic growth and long-term planning}
\end{minipage}

\begin{tcolorbox}[enhanced]
	Analyze how current economic conditions, including inflation, interest rates, consumer demand, and supply chain issues, are affecting the company and its industry. Summarize management's commentary on how these factors are influencing the business environment.
\end{tcolorbox}
\noindent\begin{minipage}{\columnwidth}
	\captionof{figure}{Question on macroeconomic impact}\label{Question on macroeconomic impact}
\end{minipage}

\begin{tcolorbox}[enhanced]
	Outline the company's cash flow management strategies and priorities discussed during the earnings call. This should include capital allocation, debt reduction, dividend policies, stock buybacks, mergers and acquisitions and investments in growth opportunities as described by management.
\end{tcolorbox}
\noindent\begin{minipage}{\columnwidth}
	\captionof{figure}{Question on financial and capital allocation management}\label{Question on financial and capital allocation management}
\end{minipage}


\input{checklist.tex}

\end{document}

%% file: checklist.tex
\section*{NeurIPS Paper Checklist}

\begin{enumerate}
	
	\item {\bf Claims}
	\item[] Question: Do the main claims made in the abstract and introduction accurately reflect the paper's contributions and scope?
	\item[] Answer: \answerYes{} 
	\item[] Justification: The abstract and introduction clearly state the main contributions.
	
	\item {\bf Limitations}
	\item[] Question: Does the paper discuss the limitations of the work performed by the authors?
	\item[] Answer: \answerYes{} 
	\item[] Justification: The paper discusses limitations in Appendix \ref{sec:limitation}.
	
	\item {\bf Theory assumptions and proofs}
	\item[] Question: For each theoretical result, does the paper provide the full set of assumptions and a complete (and correct) proof?
	\item[] Answer: \answerNA{} 
	\item[] Justification: This paper focuses on the experimental evaluation of LLMs; therefore, formal theoretical analysis is beyond its scope. To ensure rigor, we provide a detailed description of the benchmark construction.
	
	\item {\bf Experimental result reproducibility}
	\item[] Question: Does the paper fully disclose all the information needed to reproduce the main experimental results of the paper to the extent that it affects the main claims and/or conclusions of the paper (regardless of whether the code and data are provided or not)?
	\item[] Answer: \answerYes{} 
	\item[] Justification: The paper provides a detailed description of benchmark construction. Also, the benchmark is available on GitHub, which supports reproducibility.

	\item {\bf Open access to data and code}
	\item[] Question: Does the paper provide open access to the data and code, with sufficient instructions to faithfully reproduce the main experimental results, as described in supplemental material?
	\item[] Answer: \answerYes{} 
	\item[] Justification: The benchmark is available on GitHub, which supports reproducibility.

	\item {\bf Experimental setting/details}
	\item[] Question: Does the paper specify all the training and test details (e.g., data splits, hyperparameters, how they were chosen, type of optimizer) necessary to understand the results?
	\item[] Answer: \answerYes{} 
	\item[] Justification: No model training is performed in this evaluation study. For testing, we specify the datasets, task definitions, default settings, and evaluation metrics.
	
	\item {\bf Experiment statistical significance}
	\item[] Question: Does the paper report error bars suitably and correctly defined or other appropriate information about the statistical significance of the experiments?
	\item[] Answer: \answerNA{} 
	\item[] Justification: The paper does not perform statistical analysis and the stability of reported results is imporoved with the substantial number of test ECTs.
	
	\item {\bf Experiments compute resources}
	\item[] Question: For each experiment, does the paper provide sufficient information on the computer resources (type of compute workers, memory, time of execution) needed to reproduce the experiments?
	\item[] Answer: \answerYes{} 
	\item[] Justification: This paper focuses on the evaluation of proprietary state-of-the-art models. Model responses are obtained via API calls, and the downstream experiments are lightweight via laptop.
	
	\item {\bf Code of ethics}
	\item[] Question: Does the research conducted in the paper conform, in every respect, with the NeurIPS Code of Ethics \url{https://neurips.cc/public/EthicsGuidelines}?
	\item[] Answer: \answerYes{} 
	\item[] Justification: Our work is based on public data and APIs, and a further discussion on ethical considerations is available in Appendix~\ref{sec:dataset_ethics}.

	\item {\bf Broader impacts}
	\item[] Question: Does the paper discuss both potential positive societal impacts and negative societal impacts of the work performed?
	\item[] Answer: \answerNA{} 
	\item[] Justification: Our study focuses on publicly available datasets, and the tasks are confined to the content within these datasets without involving private or sensitive information. Therefore, we do not anticipate any societal impact arising from it.
	
	\item {\bf Safeguards}
	\item[] Question: Does the paper describe safeguards that have been put in place for responsible release of data or models that have a high risk for misuse (e.g., pre-trained language models, image generators, or scraped datasets)?
	\item[] Answer: \answerNA{} 
	\item[] Justification: We exclusively use publicly available datasets and API-based model access, without modifying or circumventing the built-in safety mechanisms of the LLMs.
	
	\item {\bf Licenses for existing assets}
	\item[] Question: Are the creators or original owners of assets (e.g., code, data, models), used in the paper, properly credited and are the license and terms of use explicitly mentioned and properly respected?
	\item[] Answer: \answerYes{} 
	\item[] Justification: The source datasets are public information, and the ethical considerations are provided in Appendix~\ref{sec:dataset_ethics}.
	
	\item {\bf New assets}
	\item[] Question: Are new assets introduced in the paper well documented and is the documentation provided alongside the assets?
	\item[] Answer: \answerYes{} 
	\item[] Justification: The paper introduces ECTs-100 as a new benchmark and documents its construction pipeline, datasets, and tasks.
	
	\item {\bf Crowdsourcing and research with human subjects}
	\item[] Question: For crowdsourcing experiments and research with human subjects, does the paper include the full text of instructions given to participants and screenshots, if applicable, as well as details about compensation (if any)? 
	\item[] Answer: \answerNA{} 
	\item[] Justification: The paper does not involve crowdsourcing nor research with human subjects.
	
	\item {\bf Institutional review board (IRB) approvals or equivalent for research with human subjects}
	\item[] Question: Does the paper describe potential risks incurred by study participants, whether such risks were disclosed to the subjects, and whether Institutional Review Board (IRB) approvals (or an equivalent approval/review based on the requirements of your country or institution) were obtained?
	\item[] Answer: \answerNA{} 
	\item[] Justification: The paper does not involve crowdsourcing nor research with human subjects.
	
	\item {\bf Declaration of LLM usage}
	\item[] Question: Does the paper describe the usage of LLMs if it is an important, original, or non-standard component of the core methods in this research? Note that if the LLM is used only for writing, editing, or formatting purposes and does \emph{not} impact the core methodology, scientific rigor, or originality of the research, declaration is not required.
	\item[] Answer: \answerNA{} 
	\item[] Justification: LLMs are used only for editing grammar.
	
\end{enumerate}